\documentclass{article}

\PassOptionsToPackage{numbers, compress}{natbib}
\usepackage[main,final]{neurips_2026}

\usepackage[utf8]{inputenc} %
\usepackage[T1]{fontenc}    %
\usepackage{hyperref}       %
\usepackage{url}            %
\usepackage{booktabs}       %
\usepackage{amsfonts}       %
\usepackage{nicefrac}       %
\usepackage{microtype}      %
\usepackage{xcolor}         %

\usepackage{graphicx}

\usepackage{pifont}
\newcommand{\cmark}{\ding{51}} %
\newcommand{\xmark}{\ding{55}} %
\usepackage{subcaption}
\usepackage{multirow}

\usepackage[most]{tcolorbox}
\tcbuselibrary{skins,breakable}
\usepackage{xspace}

\newcommand{\eg}{\textit{e.g.}\@\xspace}
\usepackage{wrapfig}
\usepackage{enumitem}
\usepackage{caption}
\usepackage{placeins}
\hypersetup{
  pdftitle={GeoReason: Bridging Logical Reasoning and Spatial Fidelity in Remote Sensing Segmentation},
  pdfauthor={Cheng Fan, Han Wu, Tingzhang Luo, Minjing Dong, Yehui Tang, Jianyuan Guo},
  hidelinks
}

\title{GeoReason: Bridging Logical Reasoning and Spatial Fidelity in Remote Sensing Segmentation}

\author{%
  \textbf{Cheng Fan}$^{1}$ \quad \textbf{Han Wu}$^{2}$ \quad \textbf{Tingzhang Luo}$^{1}$ \\
  \textbf{Minjing Dong}$^{1}$ \quad \textbf{Yehui Tang}$^{3}$ \quad \textbf{Jianyuan Guo}$^{1,*}$ \\[0.5em]
  $^{1}$City University of Hong Kong \\
  $^{2}$School of Computer Science, Peking University \\
  $^{3}$Samsung \\[0.4em]
  \texttt{chengfan2-c@my.cityu.edu.hk} \\
  $^{*}$Corresponding author: \texttt{jianyguo@cityu.edu.hk}
}

\begin{document}

\maketitle

\begin{abstract}
  Multimodal Large Language Models (MLLMs) have significantly advanced instruction-driven perception through the ``Embedding-as-Mask'' paradigm~\cite{lisa}. 
  However, extending this capability to Remote Sensing (RS) reasoning segmentation remains exceptionally challenging under cluttered geographical contexts and extreme scale variations, often leading to \textit{Spatial Fidelity Degradation} and \textit{Target Misidentification}.
  We identify a critical yet underexplored bottleneck: the sequential alienation of the segmentation token \texttt{[SEG]} from its initial visual grounding within the autoregressive generation paradigm.
  Produced only at the end of long reasoning chains, \texttt{[SEG]} relies on visual evidence that has been progressively diluted by linguistic abstraction and autoregressive noise. 
  This information decay is particularly catastrophic for RS targets with minuscule pixel footprints, where even marginal fidelity loss leads to localization failure.
  To address these bottlenecks, we propose \textbf{GeoReason}, a novel framework designed to fortify logical deduction and preserve spatial integrity. First, we introduce an \emph{Anticipatory Prior} mechanism that injects the \texttt{[SEG]} token directly into the initial user query, shifting it from a terminal output to a primary condition. 
  This paradigm shift transforms the token into a persistent spatial anchor, preventing information decay during long-form linguistic inference. 
  Second, we enhance visual reasoning via a \emph{Location-aware Chain-of-Thought (CoT)} that enforces coarse-to-fine localization using bounding boxes, coupled with a \emph{Latent Patch-to-Token Alignment (PTA) Loss} that explicitly aligns the latent segmentation token with physical RS textures. 
  Extensive experiments demonstrate that GeoReason consistently outperforms state-of-the-art methods on the Earthreason and RRSIS-D benchmarks.
  Code is available at \url{https://github.com/Fanc0126/GeoReason}.

\end{abstract}

\section{Introduction}
\label{sec:intro}
\begin{figure*}[t]
  \centering
  \includegraphics[width=\textwidth]{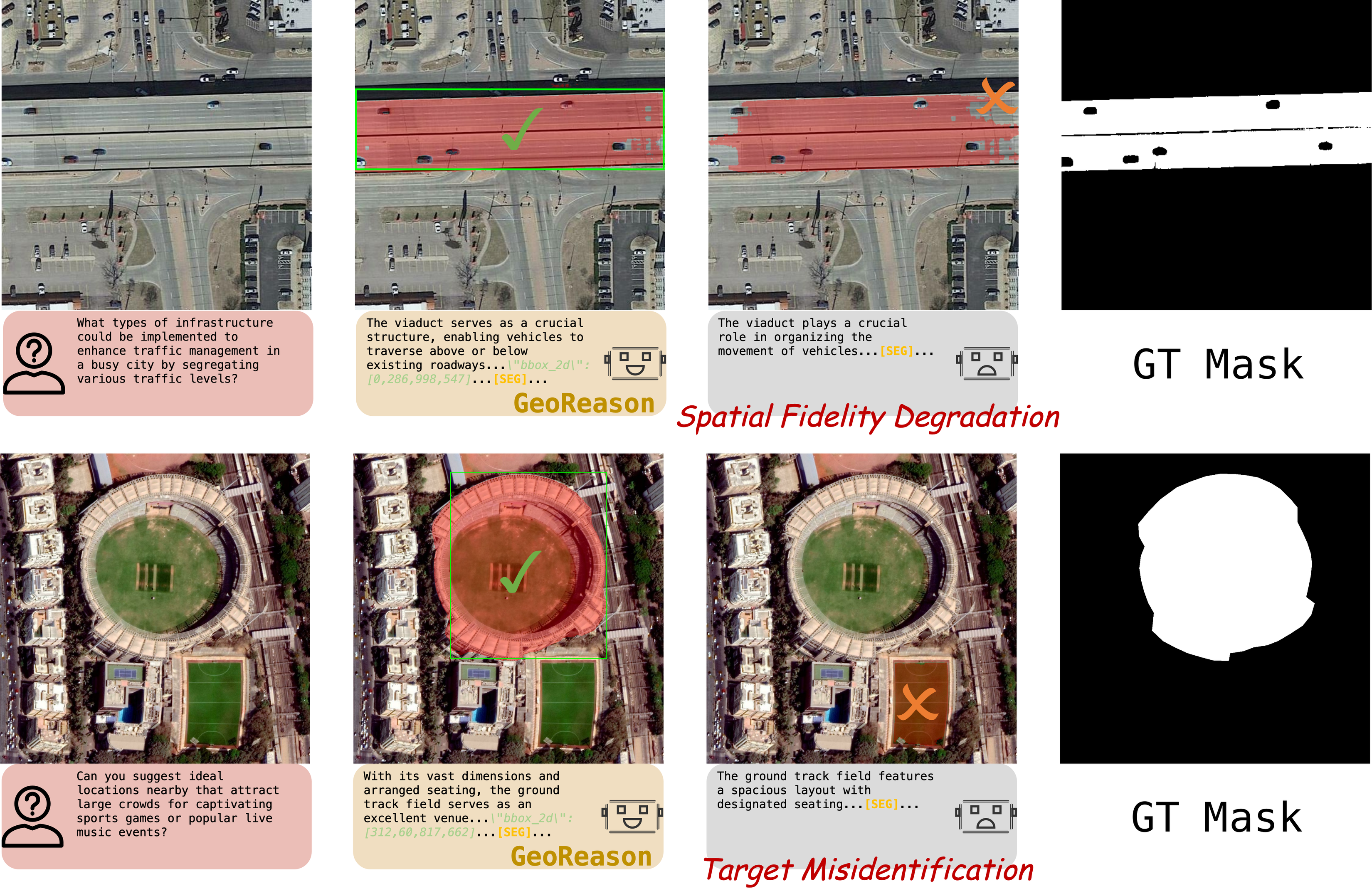}
  \vspace{-12pt}
  \caption{\small{\textbf{Qualitative comparison of GeoReason against the baseline.} 
  (Top) While the baseline identifies the viaduct, the resulting mask is fragmented and suffers from blurry boundaries.
  (Bottom) The baseline fails to interpret the functional implications of the query (``sports games or popular live music events''), leading to the segmentation of an irrelevant field.
  }}
  \vspace{-12pt}
  \label{fig:idea}
\end{figure*}

Multimodal Large Language Models (MLLMs) have shifted image segmentation from category level masks toward complex, instruction driven perception. A cornerstone of this evolution is the ``Embedding as Mask'' paradigm (\eg, LISA \cite{lisa}), which enables MLLMs to perform Reasoning Segmentation. In this framework, a dedicated token \texttt{[SEG]} is introduced into the language vocabulary and is generated autoregressively as part of the model's textual response. The hidden embedding corresponding to the generated \texttt{[SEG]} token is then extracted as a latent semantic query and projected into a mask decoder, such as the Segment Anything Model (SAM) \cite{sam}, to obtain dense pixel-level outputs. Unlike Referring Expression Segmentation (RES) which relies on explicit object descriptions, reasoning segmentation necessitates interpreting implicit, complex instructions through sophisticated logical deduction and visual grounding.

Despite its success in natural scenes, this paradigm encounters unique hurdles in Remote Sensing (RS). 
Characterized by overhead perspectives, extreme scale variations, and cluttered geographical contexts~\cite{liu2024crossmatch}, RS imagery demands not only robust logical inference to resolve functional attributes and spatial relationships, but also high-precision localization in dense environments. As illustrated in Fig.~\ref{fig:idea}, current MLLM-based RS frameworks often fall into two traps:
\begin{itemize}[leftmargin=*, itemsep=0pt, topsep=0pt]
    \item \emph{Spatial Fidelity Degradation:} The predicted mask often exhibits fragmented or imprecise boundaries despite correctly identifying the target, indicating degraded spatial consistency.
    \item \emph{Target Misidentification:} Under more severe information degradation, the segmentation may drift to an incorrect object, reflecting a failure to maintain the correct target during reasoning.
\end{itemize}

\begin{figure*}[t]
  \centering  \includegraphics[width=\textwidth]{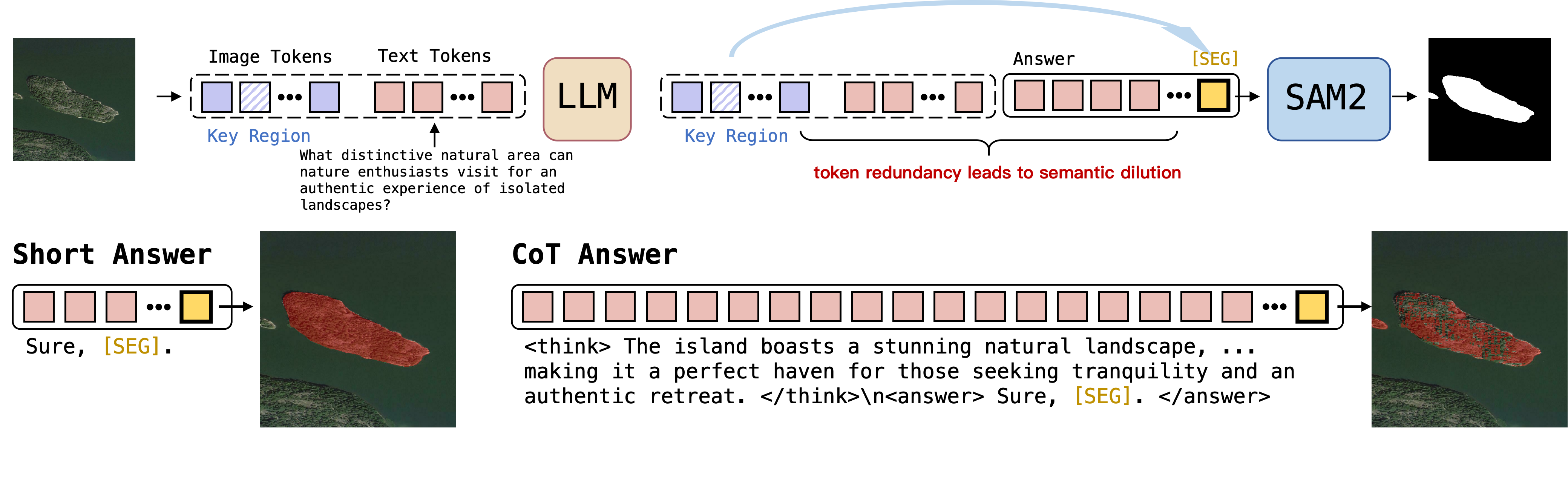}
  \vspace{-26pt}
  \caption{\small{Spatial fidelity degradation of masks induced by long reasoning chains.
  }}
  \vspace{-12pt}
  \label{fig:long_cot}
\end{figure*}

We attribute these failures to a critical yet underexplored bottleneck in current MLLM segmentation pipelines~\cite{lisa,chen2025rsrefseg,segearthr1}: the sequential alienation of the segmentation token \texttt{[SEG]} from its initial visual grounding within the autoregressive generation paradigm. In existing frameworks, \texttt{[SEG]} is produced only at the end of a long reasoning chain. As the reasoning trajectory grows longer, the hidden states of later tokens become increasingly dominated by linguistic abstraction and autoregressive noise, progressively diluting the visual evidence required for accurate spatial grounding, as illustrated in Fig.~\ref{fig:long_cot}.
Although shortening or removing the reasoning trajectory can partially alleviate this degradation, it is not a principled remedy: difficult samples often rely on multi-step visual and semantic inference, for which CoT reasoning remains essential, as detailed in Sec~\ref{sec:4.4}.
Consequently, the model may struggle to preserve precise spatial cues in the predicted mask and, in more severe cases, even lose track of the intended target during reasoning. This information decay is particularly detrimental for RS imagery, where accurate segmentation requires maintaining fine-grained structural details across cluttered geographical scenes and extreme scale variations.
To address these challenges, we propose \textbf{GeoReason} to strengthen RS-oriented visual reasoning while preserving spatial fidelity in segmentation. Our approach is built upon two core pillars:

\textbf{Anticipatory Prior via Prompt-side \texttt{[SEG]}.}
To mitigate the spatial fidelity degradation arising from the delayed generation of the segmentation token, we introduce an anticipatory prior mechanism that injects one more specialized \texttt{[SEG]} token directly into the user query. Unlike tokens generated autoregressively in the answer, this prompt-side token acts as a deterministic input that remains unaffected by the cumulative linguistic abstraction and autoregressive noise. 
By functioning as a persistent spatial anchor throughout the entire forward pass, the token can pre-extract and maintain critical visual evidence from the outset. 
This design effectively mitigates the information decay inherent in long-form RS reasoning, ensuring the spatial fidelity of the resulting latent representations.

\textbf{Enhanced Visual Reasoning for RS Domain.} To further address target misidentification in complex remote sensing scenes, we enhance the model's visual reasoning through a dual strategy that combines explicit spatial grounding with implicit latent supervision.
First, we introduce a \textit{Location-aware Chain-of-Thought (CoT)} mechanism that decouples logical deduction from final mask extraction. By incorporating bounding box prediction as an intermediate reasoning step, this design enforces a coarse-to-fine localization process, allowing the model to verify candidate regions before final mask extraction and significantly reducing the risk of selecting incorrect targets.
Complementing this explicit grounding, we propose a \textit{Latent Patch-to-Token Alignment (PTA) Loss} to directly supervise the representation of the \texttt{[SEG]} token. Specifically, we align the hidden states of this token with the physical image patches corresponding to the ground-truth target. By grounding the latent reasoning token in RS-specific textures and multi-scale geometries, this alignment strengthens its discriminative capability and ensures reliable segmentation under cluttered geographical contexts.

Our main contributions are summarized as follows:
\begin{itemize}[leftmargin=*, itemsep=0pt, topsep=0pt]
    \item We propose \textbf{GeoReason}, a novel framework for remote sensing reasoning segmentation that addresses two critical failure modes in MLLM-based segmentation: \textit{Spatial Fidelity Degradation} and \textit{Target Misidentification}.
    \item We introduce an \textit{Anticipatory Prior} mechanism that injects the \texttt{[SEG]} token directly into the query, transforming it into a persistent spatial anchor and preventing visual evidence from being diluted during long-form reasoning.
    \item We develop a \textit{Location-aware CoT} together with a \textit{Latent PTA Loss}, which jointly improve logical grounding and strengthen the spatial discriminability of the \texttt{[SEG]} token in complex RS scenes.
    \item Extensive experiments demonstrate that GeoReason achieves state-of-the-art performance on RS reasoning segmentation benchmarks, significantly improving both reasoning accuracy and segmentation quality.
\end{itemize}

\section{Related Work}

\noindent\textbf{Multimodal Large Language Models for Segmentation.} MLLM-based segmentation has moved beyond VQA toward dense, language-conditioned prediction, and existing methods mainly follow two paradigms: token-based latent extraction (Embedding-as-Mask) and text-based spatial prompt generation \cite{liu2025better,qian2025reasoning,gong2025devil,liu2025unipixel,wang2025argenseg,ying2025towards,zhu2025segagent}. LISA established the Embedding-as-Mask framework and the task of reasoning segmentation by appending a special \texttt{[SEG]} token whose hidden state prompts a mask decoder such as SAM \cite{lisa,sam,sam2,sam3}. This design has since become a standard recipe for integrating dense prediction into MLLMs \cite{xia2024gsva,rasheed2024glamm,zhang2024omg,wang2024segllm}, with extensions to multi-target reasoning in PixelLM \cite{ren2024pixellm} and to video segmentation in SA2VA and related works \cite{sa2va,yan2024visa,bai2024one,fei2024vitron,ying2025towards}. In contrast, text-only alternatives predict explicit spatial prompts such as boxes or points and then invoke an external segmentor \cite{segzero,segr1,zhu2025segagent}; representative examples include Seg-Zero \cite{segzero}, Seg-R1 \cite{segr1}, and the ``text-as-mask'' line explored by Text4Seg and Text4Seg++ \cite{text4seg,text4seg++}.

\begin{figure*}[t]
  \centering
  \includegraphics[width=\textwidth]{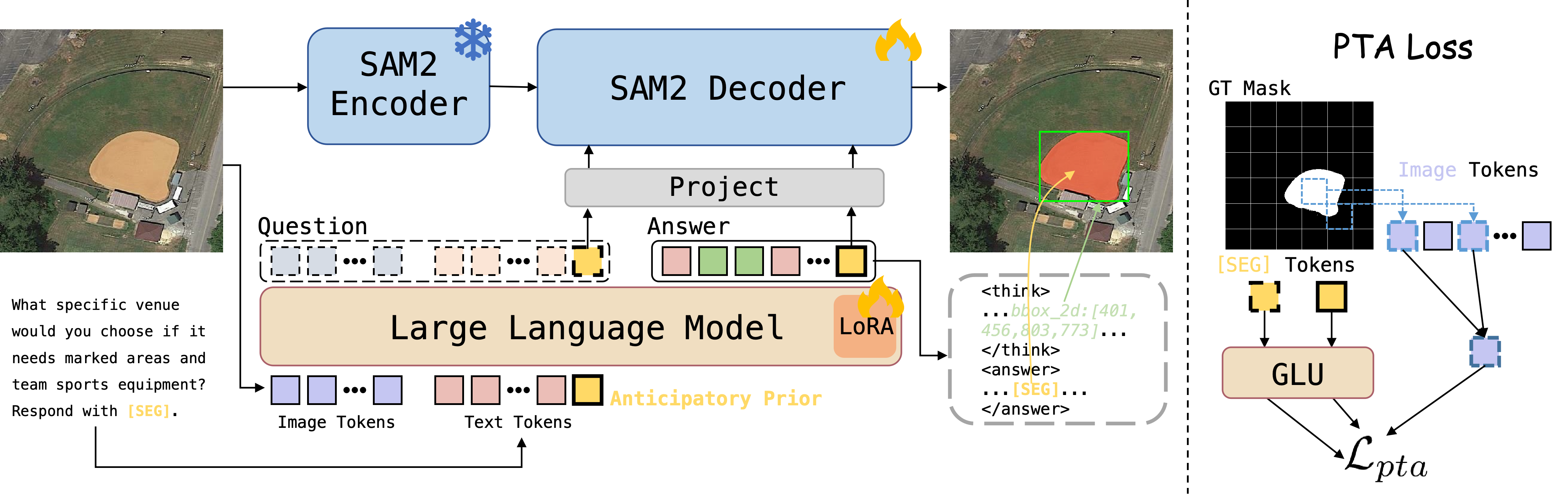}
  \vspace{-16pt}
  \caption{\small{\textbf{Overview of the GeoReason framework.} (Left) Overall architecture of \textit{Anticipatory Prior} mechanism, where a \texttt{[SEG]} token is injected into the user query to pre-anchor spatial evidence and mitigate information decay during long-chain reasoning. The MLLM is trained to output intermediate bounding boxes as spatial anchors (\textit{Location-aware CoT}) before generating the final \texttt{[SEG]}. (Right) The \textit{Latent Patch-to-Token Alignment (PTA)} loss aligns the hidden states of both prompt-side and answer-side \texttt{[SEG]} tokens with the visual patch embeddings corresponding to the GT target.}}
  \label{fig:arch}
  \vspace{-4pt}
\end{figure*}

\noindent\textbf{Language-Guided Remote Sensing Segmentation.} In remote sensing, language-guided segmentation~\cite{liu2026geoseg} has likewise evolved from referring expression segmentation~\cite{luo2026cross} to more demanding reasoning segmentation settings \cite{segearthr1,segearthr2,ni2025unigeoseg,yao2025remotereasoner}. Early RS-RES methods, such as LGCE \cite{yuan2024rrsis} and RMSIN \cite{liu2024rotated}, fuse lightweight language encoders with visual backbones, while SAM-based variants, including RS2-SAM2 \cite{rong2025rs2}, RSRefSeg \cite{chen2025rsrefseg}, and RSRefSeg2 \cite{chen2025rsrefseg2}, improve localization by translating language into sparse segmentation prompts. These models are effective for referring-style grounding but remain limited on multi-step reasoning. MLLM-based RS segmentation addresses this gap \cite{segearthr1,segearthr2}: several methods adopt the LISA-style \texttt{[SEG]} token for mask generation in referring settings \cite{ou2025geopix,shabbir2025geopixel,liu2024rsunivlm}, while SegEarth-R1 and SegEarth-R2 extend the same paradigm to the Earthreason benchmark and show that logical deduction remains difficult in cluttered RS scenes \cite{segearthr1,segearthr2}.

\noindent\textbf{Visual Reasoning in MLLMs.} CoT reasoning in MLLMs has progressed from purely textual deliberation \cite{wei2022chain,xu2025llava,wei2025perception} toward ``thinking with images'' \cite{wang2025pixel,zhang2025chain}. Recent methods strengthen reasoning with explicit visual interaction, including active perception through iterative cropping or zooming in DeepEyes, Thyme, and Pixel Reasoner \cite{zheng2025deepeyes,zhang2025thyme,wang2025pixel}, and spatial anchoring via intermediate box prediction in TreeVGR \cite{wang2025traceable}. In parallel, latent reasoning methods argue that part of the reasoning process should remain in feature space rather than text \cite{hao2024training}; for example, LVR aligns latent tokens with image representations to encode complex visual constraints \cite{li2025latent}. These trends motivate our combination of explicit spatial grounding and token-level latent supervision for RS reasoning segmentation.

\section{Our Method}
\label{sec:method}
GeoReason follows the embedding-as-mask paradigm and combines an MLLM with SAM2~\cite{sam2} for reasoning segmentation in remote sensing. The method has two main goals: preserve spatial fidelity during long-form reasoning, and improve target identification in cluttered RS scenes. To this end, we introduce a prompt-side \texttt{[SEG]} anchor (\textit{Anticipatory Prior}), a lightweight location-aware CoT, and a Latent Patch-to-Token Alignment (PTA) loss.
\subsection{System Architecture}
\label{subsec:arch}
We use Qwen3VL~\cite{bai2025qwen3} as the MLLM backbone and SAM2~\cite{sam2} as the mask decoder. Given an RS image $\mathcal{I}$ and an instruction $\mathcal{T}$, the model processes a prompt that contains a prompt-side \texttt{[SEG]} token and generates an answer that may contain one or more answer-side \texttt{[SEG]} tokens. For any \texttt{[SEG]} token, its last-layer hidden state $\mathbf{h}_{seg}\in\mathbb{R}^{d}$ is treated as a latent mask query and projected by an MLP adapter $f_{\phi}$ into the SAM2 space:
\begin{equation}
\mathcal{M}_{pred} = \mathcal{D}(\mathcal{F}_{img}, f_{\phi}(\mathbf{h}_{seg})),
\end{equation}
where $\mathcal{F}_{img}$ denotes the SAM2 image features. During training, segmentation losses are applied to both the prompt-side \texttt{[SEG]} and the answer-side \texttt{[SEG]} tokens. During inference, each of these tokens predicts a candidate mask, and the final result is obtained by taking their union. Detailed prompt formats are provided in the appendix~\ref{appendix:prompt}.

\subsection{Anticipatory Prior via Prompt-side \texttt{[SEG]} Token}
\label{subsec:prior}
In standard embedding-as-mask pipelines, the decisive \texttt{[SEG]} token is generated only after the reasoning chain, making its embedding vulnerable to attention dilution and semantic drift. This is especially problematic in RS, where fine boundaries and small structures must be preserved throughout long multimodal reasoning. We therefore place a dedicated \texttt{[SEG]} token on the prompt side, so that segmentation intent is established before autoregressive generation starts.

\begin{figure*}[t]
  \centering
  \includegraphics[width=\textwidth]{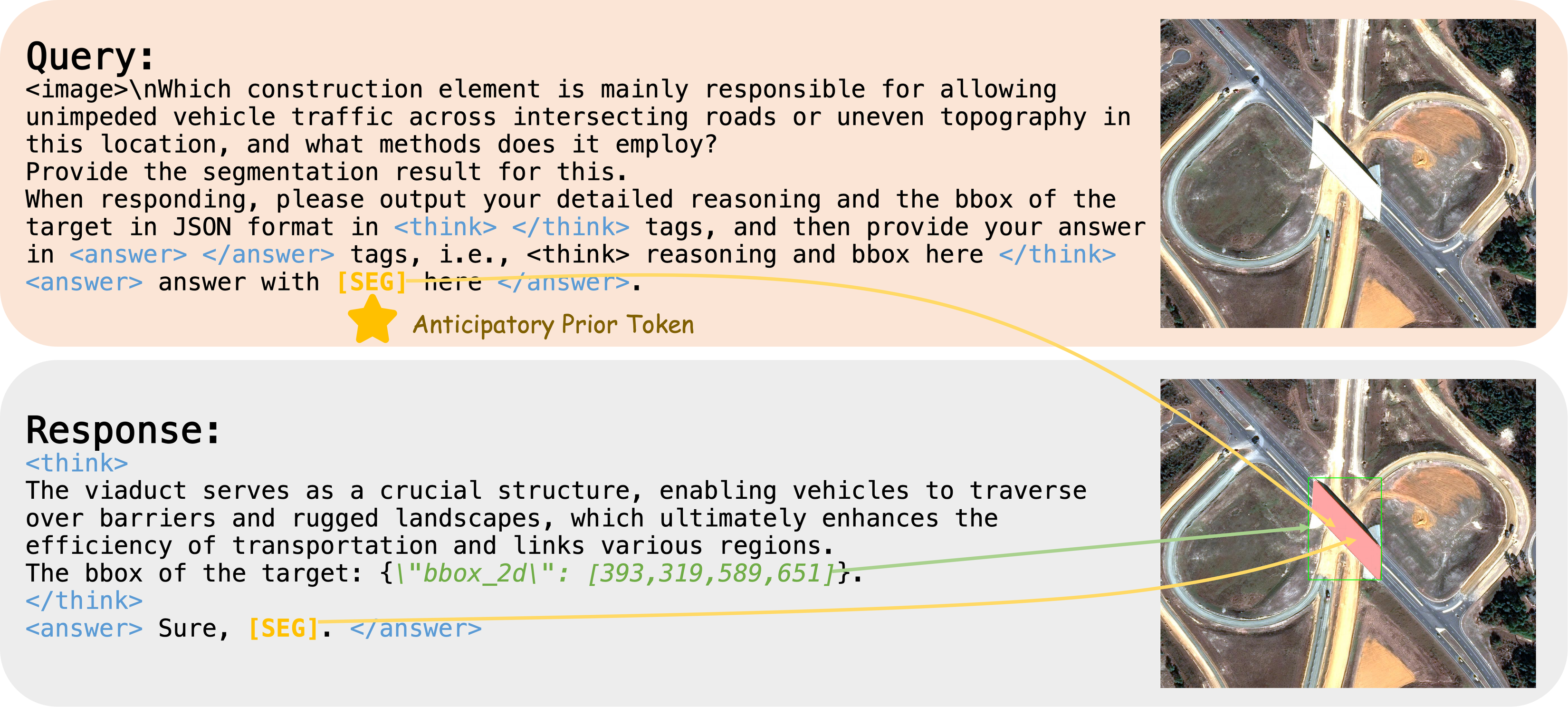}
  \caption{\small{Illustration of the prompt template and model response format. 
  }}
  \label{fig:templet}
\end{figure*}

This prompt-side token serves as an \textit{Anticipatory Prior}: it interacts with the image from the beginning of the forward pass and remains available to subsequent answer tokens as a stable spatial reference. As a result, the final answer-side \texttt{[SEG]} does not need to reconstruct segmentation cues from scratch after a long reasoning chain; instead, it can inherit a spatial prior that mitigates information decay.
\subsection{Enhanced Visual Reasoning for Remote Sensing}
\label{subsec:reasoning}
We complement the prompt-side anchor with one explicit reasoning component and one latent alignment component.

\textbf{Location-aware Chain-of-Thought.}
We ask the model to predict a bounding box $[x_{min},y_{min},x_{max},y_{max}]$ in the \texttt{<think>} stage before producing the final \texttt{[SEG]}. This design is inspired by recent think-with-images reasoning strategies, but here it plays a lightweight role: providing an explicit coarse spatial hypothesis before dense mask decoding. It is simple but useful in RS scenes with severe clutter and scale variation. In our design, BBox reasoning provides coarse localization, while the \texttt{[SEG]} embedding remains the dense latent query used by SAM2 for final mask decoding.

\textbf{Latent Vision Reasoning via PTA Loss.}
Mask supervision alone does not explicitly shape the \texttt{[SEG]} embedding to capture RS-specific textures and structures. We therefore introduce a Latent Patch-to-Token Alignment (PTA) loss that directly aligns each supervised \texttt{[SEG]} embedding with the target image patches. As shown in Fig.~\ref{fig:arch}, PTA improves the latent quality of the segmentation token rather than only its decoded mask.

Given a set of visual patch tokens $\mathcal{V} = \{\mathbf{v}_i\}_{i=1}^N$ where $\mathbf{v}_i \in \mathbb{R}^D$, let $\mathcal{I}_{valid}$ denote the index set of patches overlapping with the ground-truth mask $\mathcal{M}_{gt}$. The target visual proxy $\mathbf{z}_{vis}$ is computed via average pooling:
\begin{equation}
\mathbf{z}_{vis} = \frac{1}{|\mathcal{I}_{valid}|} \sum_{i \in \mathcal{I}_{valid}} \mathbf{v}_i.
\end{equation}
To achieve cross-modal alignment, we project the \texttt{[SEG]} token embedding $\mathbf{h}_{seg}$ into the visual representational space via a Gated Linear Unit (GLU) head $g_{\theta}(\cdot)$:
\begin{equation}
g_{\theta}(\mathbf{h}_{seg}) = \mathbf{W}_{down} \left( \sigma(\mathbf{W}_{gate} \mathbf{h}_{seg}) \odot (\mathbf{W}_{up} \mathbf{h}_{seg}) \right),
\end{equation}
where $\sigma$ denotes the SiLU activation function. The PTA loss is then formulated as the cosine distance between the projected latent embedding and the visual proxy:
\begin{equation}
\mathcal{L}_{PTA} = 1 - \frac{g_{\theta}(\mathbf{h}_{seg}) \cdot \mathbf{z}_{vis}}{\|g_{\theta}(\mathbf{h}_{seg})\| \|\mathbf{z}_{vis}\|}.
\label{eq:pta}
\end{equation}
By minimizing $\mathcal{L}_{PTA}$, the model is encouraged to encode foreground-specific visual evidence directly in the \texttt{[SEG]} embedding, which in turn improves both prompt-side anchoring and answer-side mask decoding.
In this sense, the \texttt{[SEG]} token also acts as a latent reasoning carrier, it is to make the latent representation explicitly segmentation-oriented.

\subsection{Optimization Objective}
\label{subsec:loss}
GeoReason is trained with language, mask, and latent-alignment supervision. Let $\hat{\mathcal{M}}\in\mathbb{R}^{H\times W}$ denote the predicted mask and $\mathcal{M}_{gt}$ the ground-truth mask. We use Binary Cross-Entropy and Dice losses for mask supervision:
\begin{equation}
\mathcal{L}_{bce} = -\frac{1}{HW} \sum_{i,j} \left[ (\mathcal{M}_{gt})_{i,j}\log(\hat{\mathcal{M}}_{i,j}) + \left(1-(\mathcal{M}_{gt})_{i,j}\right)\log\left(1-\hat{\mathcal{M}}_{i,j}\right) \right],
\end{equation}
\begin{equation}
\mathcal{L}_{dice} = 1 - \frac{2 \sum_{i,j} \hat{\mathcal{M}}_{i,j} \cdot (\mathcal{M}_{gt})_{i,j} + \varepsilon}{\sum_{i,j} \hat{\mathcal{M}}_{i,j}^2 + \sum_{i,j} (\mathcal{M}_{gt})_{i,j}^2 + \varepsilon}.
\end{equation}
We also apply $\mathcal{L}_{PTA}$ in Eq.~(\ref{eq:pta}). These segmentation-related losses are imposed on both the prompt-side and answer-side \texttt{[SEG]} tokens, encouraging consistent spatial semantics across the full reasoning trajectory. The overall objective is:
\begin{equation}
\mathcal{L} = \lambda_{text}\mathcal{L}_{text} + \lambda_{mask}(\mathcal{L}_{bce} + \mathcal{L}_{dice}) + \lambda_{pta}\mathcal{L}_{pta},
\end{equation}
where $\mathcal{L}_{text}$ is the standard next-token loss and $\lambda_{text},\lambda_{mask},\lambda_{pta}$ balance the three objectives.

\section{Experiments}
\subsection{Experimental Setup}
GeoReason uses Qwen3VL-4B \cite{bai2025qwen3} as the MLLM backbone and SAM2 \cite{sam2} as the mask decoder, initialized from SA2VA \cite{sa2va}. To strengthen remote-sensing priors, we further fine-tune with filtered VQA and caption samples from GeoChatInstruct \cite{kuckreja2024geochat} and VRSBench \cite{li2024vrsbench}, while removing any validation/test overlaps to avoid leakage. During training, the visual encoder is frozen, the LLM is optimized with LoRA \cite{hu2022lora}, and the MLP adapter together with SAM2 are fully updated.

We evaluate on two representative RS segmentation benchmarks: EarthReason \cite{segearthr1} for reasoning segmentation and RRSIS-D \cite{liu2024rotated} for referring segmentation. Following prior works~\cite{chen2025rsrefseg,segearthr2,lisa}, we report gIoU and cIoU. Full training hyperparameters, data preprocessing, and prompt templates are provided in the appendix~\ref{appendix:a}.

\subsection{Main Results}

\begin{table*}[tb]
\centering
\begin{minipage}[t]{0.49\linewidth}
  \centering
  \captionof{table}{\small{Comparison on EarthReason. ``Baseline'' denotes our backbone-matched post-hoc \texttt{[SEG]} model without Anticipatory Prior or Enhanced Visual Reasoning. Best results are in \textbf{bold}; second-best are \underline{underlined}.}}
  \label{tab:reasoning_results}
  \resizebox{\linewidth}{!}{
  \begin{tabular}{@{}lcccc@{}}
    \toprule
    \multirow{2}{*}{Method} & \multicolumn{2}{c}{Val} & \multicolumn{2}{c}{Test} \\
    \cmidrule(lr){2-3} \cmidrule(lr){4-5}
    & gIoU & cIoU & gIoU & cIoU \\
    \midrule
    LISA \cite{lisa} & 61.04 & 57.39 & 60.88 & 59.10 \\
    PixelLM \cite{ren2024pixellm} & 57.94 & 57.79 & 60.01 & 59.22 \\
    PSALM \cite{zhang2024psalm} & 66.61 & 62.03 & 68.30 & 64.61 \\
    SegEarth-R1 \cite{segearthr1} & 68.60 & 64.13 & 70.75 & 68.25 \\
    RemoteReasoner \cite{yao2025remotereasoner} & 69.00 & 67.80 & 71.00 & 69.10 \\
    Text4Seg++ \cite{text4seg++} & 71.90 & 69.80 & 73.00 & 65.60 \\
    SegEarth-R2 \cite{segearthr2} & 72.30 & 68.10 & 73.50 & 69.50 \\
    UniGeoSeg \cite{ni2025unigeoseg} & 72.54 & \underline{73.30} & 73.08 & \underline{74.61} \\
    \midrule
    \textit{Baseline} & \underline{75.04} & 70.33 & \underline{75.32} & 70.45 \\
    \textbf{GeoReason} & \textbf{77.61} & \textbf{77.63} & \textbf{78.61} & \textbf{80.60} \\
    \midrule
    \textit{Gain} & \textit{+2.57} & \textit{+4.33} & \textit{+3.29} & \textit{+5.99} \\
    \bottomrule
  \end{tabular}
  }
\end{minipage}
\hfill
\begin{minipage}[t]{0.49\linewidth}
  \centering
  \captionof{table}{\small{Comparison on RRSIS-D. ``-'' indicates that the metric was not reported by the original authors.}}
  \label{tab:rrsisd_results}
  \resizebox{\linewidth}{!}{
  \begin{tabular}{@{}lcccc@{}}
    \toprule
    \multirow{2}{*}{RRSIS-D} & \multicolumn{2}{c}{Val} & \multicolumn{2}{c}{Test} \\
    \cmidrule(lr){2-3} \cmidrule(lr){4-5}
    & gIoU & cIoU & gIoU & cIoU \\
    \midrule
    \textit{Traditional Methods} & & & & \\
    LGCE \cite{yuan2024rrsis} & 60.16 & 76.68 & 59.37 & 76.34 \\
    RIS-DMMI \cite{hu2023beyond} & 60.72 & 77.01 & 60.12 & 76.20 \\
    LAVT \cite{yang2022lavt} & 61.46 & 77.59 & 61.04 & 77.19 \\
    RMSIN \cite{liu2024rotated} & 65.10 & 78.27 & 64.20 & 77.79 \\
    \midrule
    \textit{MLLM-based Methods} & & & & \\
    LISA \cite{lisa} & 27.84 & - & 26.78 & - \\
    PixelLM \cite{ren2024pixellm} & 33.89 & - & 31.65 & - \\
    NEXT-Chat \cite{zhang2023next} & 26.98 & - & 24.98 & - \\
    GeoGround \cite{zhou2024geoground} & 61.10 & - & 60.50 & - \\
    Text4Seg++ \cite{text4seg++} & 64.10 & 75.80 & 62.80 & 74.20 \\
    SegEarth-R1 \cite{segearthr1} & 67.56 & 78.92 & 66.40 & 78.01 \\
    GeoPixel \cite{shabbir2025geopixel} & \underline{67.99} & \textbf{81.77} & \underline{67.30} & \textbf{84.90} \\
    \midrule
    \textbf{GeoReason} & \textbf{68.36} & \underline{79.87} & \textbf{67.32} & \underline{79.39} \\
    \bottomrule
  \end{tabular}
  }
\end{minipage}
\end{table*}

Tables \ref{tab:reasoning_results} and \ref{tab:rrsisd_results} show that GeoReason performs strongly on both reasoning and referring segmentation, with the largest gains appearing in the reasoning setting that motivates our design.

\textbf{Reasoning Segmentation.} GeoReason establishes a new SOTA on EarthReason across all metrics. The runner-up differs by metric: our backbone-matched baseline is second-best on gIoU, while UniGeoSeg is second-best on cIoU. Relative to these strongest alternatives, GeoReason improves val/test gIoU by $+2.57/+3.29$ and val/test cIoU by $+4.33/+5.99$. Notably, the baseline already benefits from a strong Qwen3VL+SAM2 backbone, yet still trails GeoReason by a large margin, especially on test cIoU ($80.60\%$ vs.\ $70.45\%$). This shows that the gains do not come from backbone strength alone, but from stabilizing the segmentation token during long-form reasoning and improving its spatial grounding.

\textbf{Referring Segmentation.} On RRSIS-D, GeoReason achieves the best gIoU on both validation and test, and remains competitive in cIoU against specialized referring models. Since this benchmark involves little explicit reasoning, the gains mainly reflect better mask quality. This is consistent with the role of PTA: beyond helping reasoning segmentation, it also improves the discriminability of the \texttt{[SEG]} representation for standard referring scenarios.

\subsection{Mechanistic Analysis of the Anticipatory Prior}
\label{sec:4.3}
\begin{figure}[t]
  \centering
  \begin{subfigure}[t]{0.46\linewidth}
    \centering
    \includegraphics[width=0.96\linewidth]{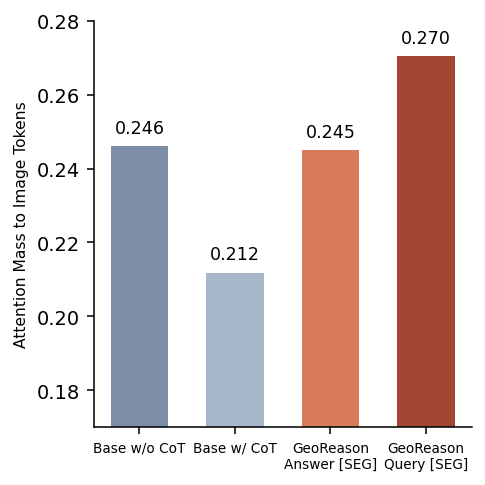}
    \caption{\small{Direct attention from different \texttt{[SEG]} tokens to image tokens, averaged over the last four layers.}}
    \label{fig:seg_img_attn}
  \end{subfigure}
  \hfill
  \begin{subfigure}[t]{0.46\linewidth}
    \centering
    \includegraphics[width=0.96\linewidth]{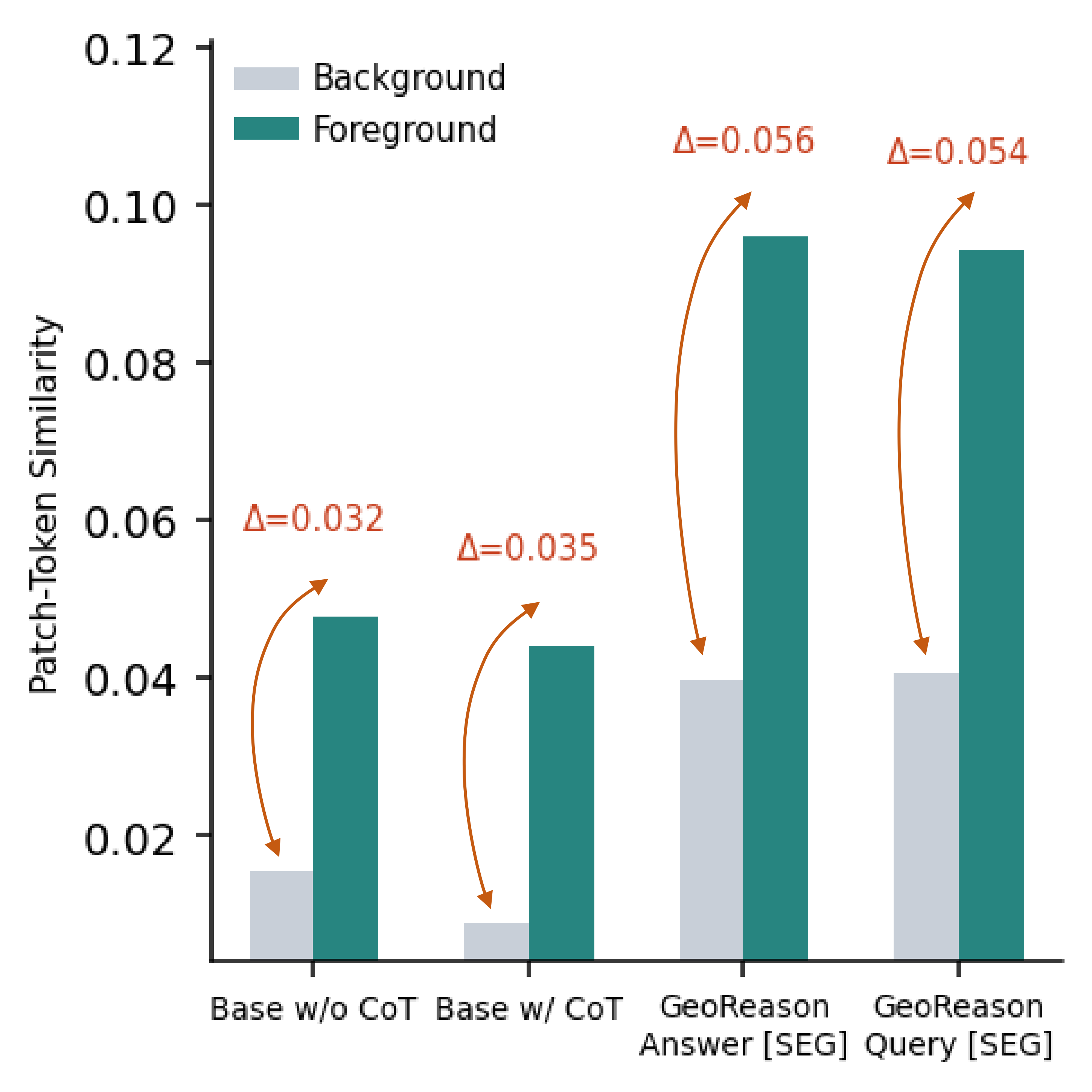}
    \caption{\small{Foreground/background patch alignment and token-only mask quality of different \texttt{[SEG]} tokens.}}
    \label{fig:seg_patch_align}
  \end{subfigure}
  \caption{\small{\textbf{Token-level mechanism analysis.} (a) Direct attention from different \texttt{[SEG]} token sources to image tokens. (b) Foreground/background alignment and token-only mask quality.}}
  \label{fig:seg_mechanism_stats}
\end{figure}

\begin{figure}[t]
  \centering
  \begin{subfigure}[t]{0.62\linewidth}
    \vspace{0pt}
    \centering
    \includegraphics[width=\linewidth]{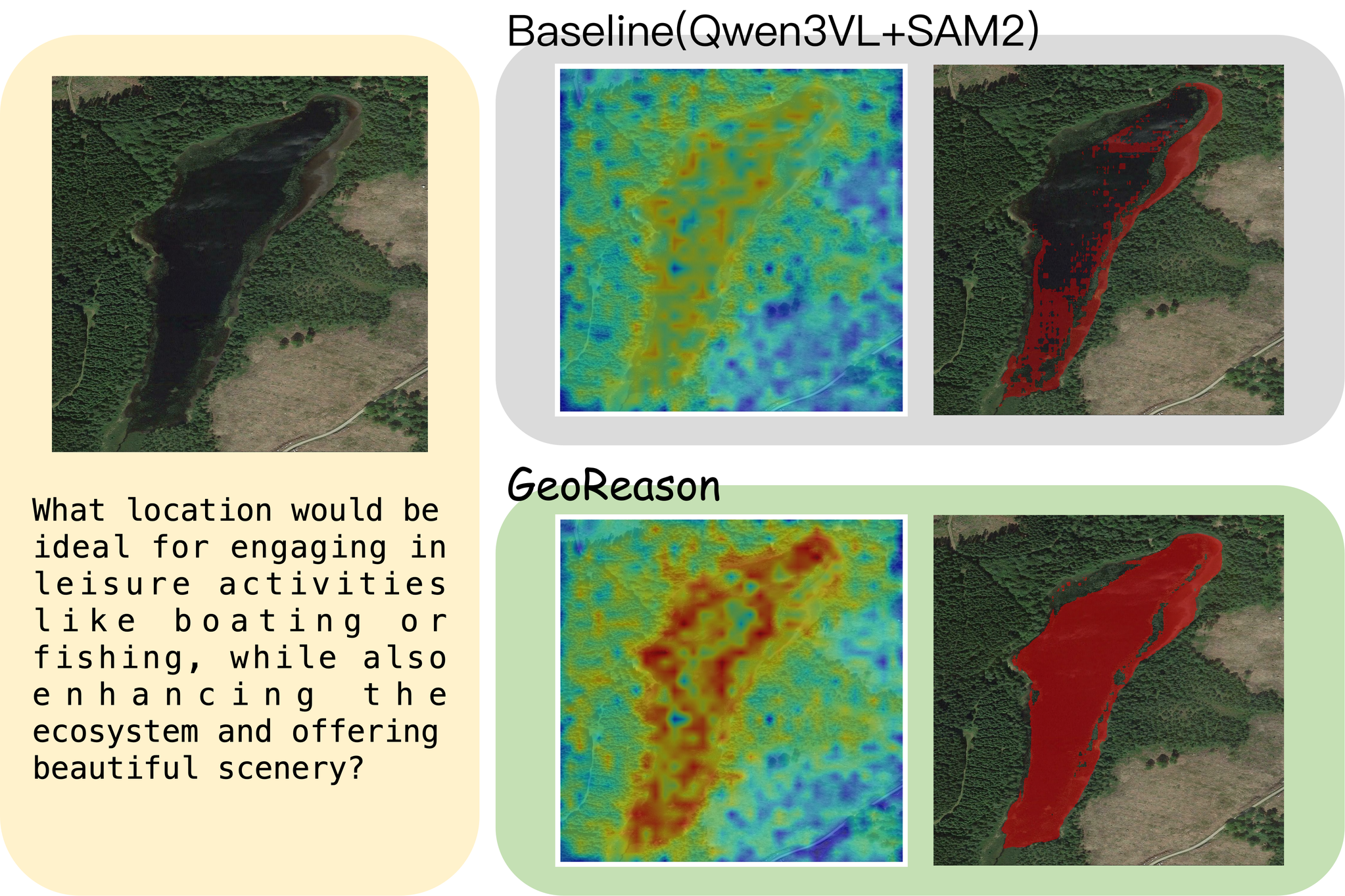}
    \caption{\small{Qualitative comparison on a representative case.}}
    \label{fig:seg_case_study}
  \end{subfigure}
  \hfill
  \begin{subfigure}[t]{0.36\linewidth}
    \vspace{0pt}
    \centering
    \includegraphics[width=0.9\linewidth]{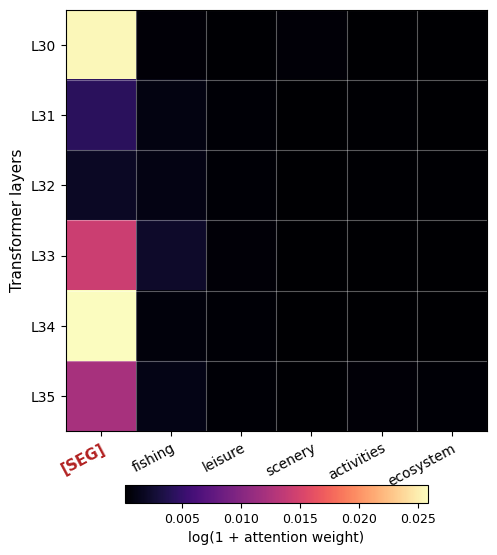}
    \vspace{-6pt}
    \caption{\scriptsize{
    Late-layer attention distribution of the \textbf{final answer-side} \texttt{[SEG]} token over the prompt-side \texttt{[SEG]} token and selected key words, with other tokens omitted for clarity.}}
    \label{fig:seg_query_heatmap}
  \end{subfigure}
  \caption{\small{\textbf{Case-level evidence.} (a) GeoReason preserves focus on the target region and produces a cleaner mask than the baseline. (b) In the same case, the final answer-side \texttt{[SEG]} focuses strongly on the prompt-side \texttt{[SEG]}.}}
  \label{fig:seg_mechanism_case}
\end{figure}

\begin{table}[tb]
  \captionsetup{position=top,skip=6pt}
  \caption{\small{Ablation study of GeoReason components. AP: Anticipatory Prior, BBox: Location-aware CoT, PTA: Latent PTA Loss. g: gIoU, c: cIoU (\%).}}
  \label{tab:ablation_complete}
  \centering
  \small 
  \setlength{\tabcolsep}{3pt} 
  \begin{tabular}{@{}ccc|cc|cc|cc|cc@{}}
    \toprule
    \multicolumn{3}{c|}{Components} & \multicolumn{4}{c|}{Earthreason (Reasoning)} & \multicolumn{4}{c}{RRSIS-D (Referring)} \\
    \cmidrule(lr){1-3} \cmidrule(lr){4-7} \cmidrule(lr){8-11}
    AP & BBox & PTA & Val g & Val c & Test g & Test c & Val g & Val c & Test g & Test c \\ 
    \midrule
    \xmark & \xmark & \xmark & 75.04 & 70.33 & 75.32 & 70.45 & 67.13 & 79.39 & 66.42 & 78.67 \\
    \xmark & \xmark & \cmark & 76.02 & 72.26 & 76.32 & 73.01 & 67.98 & 79.71 & \underline{66.99} & \underline{79.32} \\
    \xmark & \cmark & \xmark & 74.42 & 70.89 & 74.54 & 70.59 & 65.60 & 76.13 & 65.29 & 76.43 \\
    \cmark & \xmark & \xmark & 76.24 & \underline{77.32} & \underline{78.02} & \underline{78.87} & 67.54 & 79.45 & 66.13 & 78.32 \\
    \midrule
    \cmark & \xmark & \cmark & 76.45 & 75.64 & 77.39 & 77.18 & \underline{68.12} & \underline{79.88} & 66.32 & 78.67 \\
    \cmark & \cmark & \xmark & \underline{76.66} & 76.24 & 77.25 & 76.62 & 67.58 & \textbf{79.93} & 66.27 & 78.28 \\
    \midrule
    \cmark & \cmark & \cmark & \textbf{77.61} & \textbf{77.63} & \textbf{78.61} & \textbf{80.60} & \textbf{68.36} & 79.87 & \textbf{67.32} & \textbf{79.39} \\
    \bottomrule
  \end{tabular}
\end{table}
To validate the role of the Anticipatory Prior, we compare four token sources on the EarthReason test set: the baseline answer-side \texttt{[SEG]} without CoT, the baseline answer-side \texttt{[SEG]} with CoT, the GeoReason answer-side \texttt{[SEG]}, and the GeoReason prompt-side \texttt{[SEG]}. Fig.~\ref{fig:seg_mechanism_stats}\subref{fig:seg_img_attn} shows that long autoregressive reasoning weakens direct visual access in the baseline: attention mass from the final \texttt{[SEG]} to image tokens drops from $0.246$ without CoT to $0.212$ with CoT. GeoReason largely removes this degradation. Its prompt-side \texttt{[SEG]} has the strongest image attention ($0.270$), and its answer-side \texttt{[SEG]} retains $0.245$, nearly recovering the no-CoT level despite a longer reasoning sequence. This is direct evidence that the prompt-side token acts as a persistent spatial anchor rather than an extra symbolic token.

We further ask whether the final answer-side \texttt{[SEG]} explicitly uses this anchor. After removing image and template tokens, we rank prompt tokens by their mean attention from the answer-side \texttt{[SEG]} over the last four layers. The prompt-side \texttt{[SEG]} has mean rank $2.01$, appears in the top-3 attended prompt tokens for $99.69\%$ of samples, and in the top-5 for $99.95\%$. The case heatmap in Fig.~\ref{fig:seg_mechanism_case}\subref{fig:seg_query_heatmap} shows the same behavior at the instance level: over the last six layers, the answer-side \texttt{[SEG]} attends much more strongly to the prompt-side \texttt{[SEG]} than to content words such as \textit{fishing}, \textit{leisure}, and \textit{scenery}. Fig.~\ref{fig:seg_mechanism_stats}\subref{fig:seg_patch_align} further shows that GeoReason also improves segmentation-oriented alignment: relative to the baseline CoT answer-side \texttt{[SEG]}, the GeoReason answer-side \texttt{[SEG]} raises foreground similarity from $0.044$ to $0.096$ and enlarges the foreground-background margin from $0.035$ to $0.056$. The qualitative case in Fig.~\ref{fig:seg_mechanism_case}\subref{fig:seg_case_study} is consistent with these statistics, with GeoReason producing a more complete lake mask. Together, these results support our claim that AP preserves visual evidence while PTA makes the resulting \texttt{[SEG]} representation more discriminative for the target region.

\subsection{Ablation Studies}
\label{sec:4.4}
Here we conduct ablation studies to disentangle the roles of the Anticipatory Prior (AP), the Location-aware CoT (BBox), and the Latent PTA loss (PTA). Table~\ref{tab:ablation_complete} reports the core component ablations on EarthReason and RRSIS-D, and Table~\ref{tab:discussion_complete} provides two targeted diagnostic variants.

\begin{table*}[tb]
\centering
\begin{minipage}[t]{0.49\linewidth}
  \captionof{table}{\small{Diagnostic studies on EarthReason. \textbf{Shifted-SEG}: move the prompt-side \texttt{[SEG]} into the answer CoT. \textbf{Zero-Reasoning}: a short-answer baseline trained and evaluated without CoT. \textbf{Avg} is the mean of gIoU and cIoU across Val and Test.}}
  \label{tab:discussion_complete}
\end{minipage}\hfill
\begin{minipage}[t]{0.49\linewidth}
  \captionof{table}{\small{Oracle-referring diagnostics on EarthReason. \textbf{Oracle Ref.}: convert each reasoning query into a direct referring expression using Qwen3-30B. \textbf{Oracle Ref. + PTA}: additionally apply PTA in this simplified setting. \textbf{Avg} is the mean of gIoU and cIoU across Val and Test.}}
  \label{tab:oracle_referring}
\end{minipage}
\par\vspace{6pt}
\begin{minipage}[t]{0.49\linewidth}
  \centering
  \footnotesize
  \renewcommand{\arraystretch}{1.05}
  \setlength{\tabcolsep}{1pt}
  \begin{tabular*}{\linewidth}{@{\extracolsep{\fill}}l|cccc|c@{}}
    \toprule
    \multirow{2}{*}{Setting} & \multicolumn{2}{c}{Val} & \multicolumn{2}{c|}{Test} & \multirow{2}{*}{Avg} \\
    \cmidrule(lr){2-3} \cmidrule(lr){4-5}
    & gIoU & cIoU & gIoU & cIoU & \\
    \midrule
    Baseline & 75.04 & 70.33 & 75.32 & 70.45 & 72.79 \\
    GeoReason & 77.61 & 77.63 & 78.61 & 80.60 & 78.61 \\
    \midrule
    Shifted-SEG & 75.65 & 73.60 & 76.22 & 71.07 & 74.14 \\
    Zero-Reasoning & 75.64 & 72.59 & 72.86 & 77.43 & 74.63 \\
    \bottomrule
  \end{tabular*}
\end{minipage}
\hfill
\begin{minipage}[t]{0.49\linewidth}
  \centering
  \footnotesize
  \renewcommand{\arraystretch}{1.05}
  \setlength{\tabcolsep}{1pt}
  \begin{tabular*}{\linewidth}{@{\extracolsep{\fill}}l|cccc|c@{}}
    \toprule
    \multirow{2}{*}{Setting} & \multicolumn{2}{c}{Val} & \multicolumn{2}{c|}{Test} & \multirow{2}{*}{Avg} \\
    \cmidrule(lr){2-3} \cmidrule(lr){4-5}
    & gIoU & cIoU & gIoU & cIoU & \\
    \midrule
    Baseline & 75.04 & 70.33 & 75.32 & 70.45 & 72.79 \\
    GeoReason & 77.61 & 77.63 & 78.61 & 80.60 & 78.61 \\
    \midrule
    Oracle Ref. & 79.20 & 82.16 & 79.34 & 82.67 & 80.84 \\
    Oracle Ref. + PTA & 79.94 & 82.70 & 80.33 & 84.14 & 81.78 \\
    \bottomrule
  \end{tabular*}
\end{minipage}
\end{table*}
\textbf{Component contributions.} AP is the dominant factor on reasoning segmentation: adding AP alone improves EarthReason test cIoU from $70.45\%$ to $78.87\%$, confirming that early spatial anchoring is critical under autoregressive reasoning. PTA alone also gives a clear gain ($70.45\%\rightarrow73.01\%$) and is the main source of improvement on RRSIS-D, where mask quality matters more than long-form reasoning. By contrast, BBox alone is slightly below the baseline on EarthReason, indicating that longer reasoning does not help unless the segmentation token is first stabilized.

\textbf{Complementarity.} The full model performs best on both benchmarks. Partial variants such as AP+PTA or AP+BBox remain competitive but consistently trail the full design on EarthReason, showing that the three modules address different error sources: AP preserves spatial evidence, BBox improves coarse localization, and PTA sharpens the latent segmentation representation.

\begin{figure}[!htbp]
  \centering
  \includegraphics[width=0.98\linewidth]{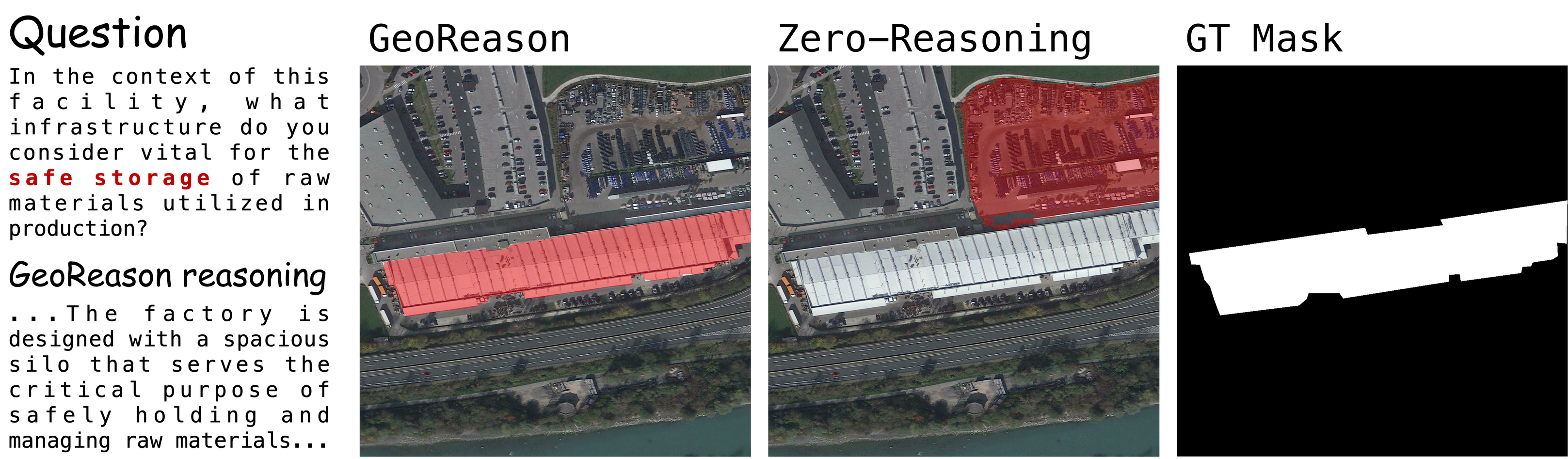}
  \vspace{-0.5em}
  \caption{\small{\textbf{Reasoning-sensitive case study on EarthReason.} GeoReason correctly selects the protected storage structure required for \emph{safe} raw-material storage, whereas the Zero-Reasoning variant overfits to a visually plausible but functionally incorrect open yard.}}
  \label{fig:reason_case}
  \vspace{-0.8em}
\end{figure}

\textbf{Diagnostic variants.} \textit{Shifted-SEG} keeps the additional \texttt{[SEG]} token but moves it from the prompt into the answer stream, reducing the average score to $74.14\%$. This confirms that the improvement does not come from simply adding another special token; what matters is prompt-side placement, which allows the token to condition the full forward pass. \textit{Zero-Reasoning}, a short-answer variant without CoT, outperforms the full reasoning baseline ($74.63\%$ vs.\ $72.79\%$ Avg), indicating that CoT can hurt segmentation if no spatial-preservation mechanism is provided. However, it still remains well below full GeoReason ($78.61\%$ Avg), showing that the goal is not to remove reasoning, but to make reasoning compatible with spatial fidelity. Fig.~\ref{fig:reason_case} gives a example: for the query \emph{``In the context of this facility, what infrastructure do you consider vital for the safe storage of raw materials utilized in production?''}, GeoReason reasons that the protected silo is the correct target, whereas the Zero-Reasoning model drifts to the exposed open area where materials are merely placed.

\textbf{Quantifying the reasoning burden in RS.} To verify that reasoning itself is a major bottleneck, we construct an \textit{Oracle Referring} setting by rewriting each EarthReason query into a direct referring expression with Qwen3-30B \cite{yang2025qwen3}, thereby preserving the target semantics while removing the need for autonomous logical deduction. This converts the task from reasoning segmentation into a simplified referring-style segmentation problem and thus approximates the segmentation ceiling once reasoning difficulty is stripped away. As shown in Table~\ref{tab:oracle_referring}, the baseline reasoning model ($72.79\%$ Avg) is 8.05 below Oracle Referring ($80.84\%$ Avg), quantifying a substantial \emph{reasoning cost} in RS segmentation. Importantly, GeoReason reaches $78.61\%$ Avg, recovering 5.82 of this gap and leaving only a 2.23 residual to the oracle ceiling. This supports our central claim: beyond preserving spatial fidelity, GeoReason also strengthens reasoning itself, bringing complex reasoning segmentation substantially closer to direct referring segmentation. We further apply PTA on top of Oracle Referring. Even after the reasoning burden is removed, PTA still raises performance to $81.78\%$ Avg and $84.14\%$ test cIoU. This mirrors the gains on RRSIS-D and indicates that PTA is not merely compensating for reasoning errors; it acts as a task-agnostic enhancer of the \texttt{[SEG]} token by improving its latent alignment with RS textures and structures.

\FloatBarrier
\section{Conclusion}
We presented \textbf{GeoReason}, a framework for remote-sensing reasoning segmentation that addresses \textit{Spatial Fidelity Degradation} and \textit{Target Misidentification}. By combining a prompt-side \texttt{[SEG]} anchor, Location-aware CoT, and a PTA loss, GeoReason preserves visual evidence, improves grounding, and strengthens the segmentation-token representation. Experiments on EarthReason and RRSIS-D demonstrate state-of-the-art performance on both reasoning and referring segmentation.

\clearpage
\bibliographystyle{plainnat}
\bibliography{main}
\clearpage
\appendix
\raggedbottom

\section{Implementation Details}
\label{appendix:a}
In this part, we provide additional details regarding the experimental setup, including data composition, hyperparameter configurations, and the prompt templates used for training and inference.

\subsection{Data Composition and Preprocessing}
To endow GeoReason with foundational geographical cognition and robust understanding of remote sensing (RS) semantic attributes, we incorporate auxiliary data from \textbf{GeoChatInstruct} \cite{kuckreja2024geochat} and \textbf{VRSBench} \cite{li2024vrsbench} during the training phase. These datasets are large-scale RS benchmarks covering diverse tasks. To maintain a focus on general RS perception, we exclusively extract the Visual Question Answering (VQA) and Captioning subsets.

\paragraph{Data Leakage Prevention.} A significant portion of images in GeoChatInstruct and VRSBench originate from the DIOR dataset \cite{li2020object}, which is also the source for the RRSIS-D \cite{yuan2024rrsis} benchmark. To ensure the integrity of our evaluation and prevent data leakage, we implemented a strict data cleansing process. We identified and removed all images (and their corresponding crops) from the auxiliary sets that overlap with the validation or test sets of RRSIS-D. After filtering, we obtained \textbf{21.2k} samples from GeoChatInstruct and \textbf{34.7k} samples from VRSBench. During training, these samples are interleaved with the segmentation datasets. Notably, these auxiliary tasks do not involve the \texttt{[SEG]} token or mask generation.

\subsection{Training Configurations}
\paragraph{Optimization.} We employ the AdamW \cite{loshchilov2017decoupled} optimizer with an initial learning rate of $4 \times 10^{-5}$ and a weight decay of $0.05$. A linear warmup strategy is applied for the first $5\%$ of iterations. To ensure training stability, we use gradient clipping with a maximum norm of $1.0$. 

\paragraph{Model Fine-tuning.} The vision encoder of the MLLM remains frozen throughout the training process. The MLP adapter and the SAM2 mask decoder are fully updated. For the Large Language Model (LLM), we utilize LoRA with a rank $r=128$, $\alpha=256$, and a dropout rate of $0.05$. The maximum sequence length is set to 8192. Regarding the multi-task optimization, the balancing coefficients $\lambda_{text}$, $\lambda_{mask}$, and $\lambda_{pta}$ are all set to 1.0 in our implementation.

\paragraph{Hardware and Efficiency.} The model is trained for a single epoch with a total effective batch size of $64$. We utilize $4 \times$ NVIDIA RTX Pro 6000 GPUs, employing a per-GPU batch size of $2$ combined with $8$-step gradient accumulation. The training is conducted using BF16 mixed-precision and takes approximately $9$ hours to complete. Benchmarked on the EarthReason validation set, GeoReason increases latency ($1223.6$ ms vs. $774.1$ ms) and token count ($73.3$ vs. $45.1$) compared to the baseline (Qwen3VL-4B + SAM2 without our proposed methods). This overhead stems primarily from the autoregressive generation of reasoning and bounding box tokens rather than model complexity, representing a highly favorable trade-off for the significant performance gains.

\subsection{Prompt Templates and Formats}
\label{appendix:prompt}
We define specific prompt templates for Reasoning Segmentation (RS) and Referring Expression Segmentation (RES). The bounding box coordinates used in the \texttt{<think>} tags are derived from the ground-truth masks by calculating the minimum axis-aligned box that covers all target pixels.

\begin{tcolorbox}[
    colback=teal!5,          %
    colframe=teal!60!black,  %
    arc=4pt, 
    boxrule=0.6pt, 
    title=\textbf{Prompt Templates for GeoReason Training and Inference}, 
    coltitle=white,          %
    fonttitle=\bfseries\small, 
    left=8pt, right=8pt, top=8pt, bottom=8pt
]
\small

\textbf{Task A: Reasoning Segmentation} \\
\textit{Query Template:} \\
\texttt{<image> [Question] When responding, please output your detailed reasoning and the bbox of the target in JSON format in <think> </think> tags, and then provide your answer in <answer> </answer> tags, i.e., <think> reasoning and bbox here </think> <answer> answer with [SEG] here </answer>.}

\vspace{0.4em}
\textit{Model Response Format:} \\
\texttt{<think> [Detailed Reasoning Content] The bbox of the target: \{"bbox\_2d": [xmin, ymin, xmax, ymax]\} </think> \textbackslash n <answer> [Answer Content] </answer>.}

\vspace{0.8em}
\hrule %
\vspace{0.8em}

\textbf{Task B: Referring Expression Segmentation} \\
\textit{Query Template:} \\
\texttt{<image> [Question] When responding, please output the bbox of the target in JSON format in <think> </think> tags, and then provide your answer in <answer> </answer> tags, i.e., <think> bbox here </think> <answer> answer with [SEG] here </answer>.}

\vspace{0.4em}
\textit{Model Response Format:} \\
\texttt{<think> The bbox of the target: \{"bbox\_2d": [xmin, ymin, xmax, ymax]\} </think> \textbackslash n <answer> [Answer Content] </answer>.}

\vspace{0.4em}
\begin{tcolorbox}[
    colback=white, 
    colframe=teal!30,        %
    boxrule=0.4pt, 
    arc=2pt, 
    left=4pt, right=4pt, top=4pt, bottom=4pt
]
    \footnotesize
    \textbf{Note:} During training, both the \texttt{[SEG]} token in the query and the \texttt{[SEG]} token(s) in the \texttt{<answer>} tag are jointly supervised using $\mathcal{L}_{bce}, \mathcal{L}_{dice},$ and $\mathcal{L}_{pta}$.
\end{tcolorbox}

\end{tcolorbox}

\paragraph{Supervision and Inference.} During training, our segmentation-related loss is applied to both the prompt-side \texttt{[SEG]} token and any \texttt{[SEG]} tokens generated within the \texttt{<answer>} tags. During inference, we extract the latent embeddings of the prompt-side \texttt{[SEG]} token and all \texttt{[SEG]} tokens present in the generated answer. Each token produces a candidate mask through the SAM2 decoder, and the final segmentation result is obtained by taking the union of all predicted masks.

\section{Visual-Evidence Dilution in Long Reasoning}
\label{appendix:controlled_length}

\subsection{Analysis of Visual-Evidence Dilution}
During visually grounded tasks, extended textual reasoning can weaken multimodal models' reliance on visual evidence in subsequent answers. Several studies have examined this phenomenon. Sun et al.~\cite{sun2025visualforgetting} analyze long-CoT \emph{visual forgetting}: removing the image midway through reasoning changes MathVista-Hard accuracy only from $43.1\%$ to $40.9\%$, while attention to visual tokens declines as reasoning proceeds. Tian et al.~\cite{tian2026morethought} use early decisions to examine reasoning progress: MMStar and HallusionBench accuracy drops by over two percentage points after its peak; among Vision-R1's full-reasoning errors, $55.23\%$ are perception errors, and $32.35\%$ of these can be corrected at an earlier decision point. MME-CoT~\cite{jiang2025mmecot} reports negative stability when switching from direct answers to CoT on perception tasks for most models, with a maximum drop of $6.8$ percentage points. OPERA~\cite{huang2024opera} observes that longer generation can cause later tokens to over-trust nearby summary tokens and neglect earlier image tokens.

In the embedding-as-mask setting, longer reasoning may dilute visual evidence through two mechanisms: \textbf{(i) visual-token competition}, where additional reasoning tokens compete with image tokens for attention from the answer-side \texttt{[SEG]}, and \textbf{(ii) mask-query drift}, where the evolving reasoning context shifts its hidden representation away from mask-relevant visual evidence. We analyze these mechanisms through image-attention mass and segmentation-query representation drift.

Let $\mathcal{R}_L$ be a reasoning prefix of $L$ tokens, $\mathcal{I}_{img}$ the image-token indices, and $\mathcal{C}_L$ the context visible to the final answer-side \texttt{[SEG]}. For one attention head at layer $\ell$, its image-attention mass is
\begin{equation}
A_{img}^{(\ell)}(L)=
\sum_{i\in\mathcal{I}_{img}}
\frac{\exp\!\left((\mathbf{q}_{seg}^{(\ell)}(L))^\top\mathbf{k}_{i}^{(\ell)}/\sqrt{d}\right)}
{\sum_{j\in\mathcal{C}_L}\exp\!\left((\mathbf{q}_{seg}^{(\ell)}(L))^\top\mathbf{k}_{j}^{(\ell)}/\sqrt{d}\right)},
\label{eq:prefix_visual_attention}
\end{equation}
where $d$ is the head dimension. Additional reasoning tokens enlarge the competing linguistic context, while the segmentation query and attention logits can also change.

Using the notation of Sec.~\ref{subsec:arch}, the answer-side mask after this prefix is
\begin{equation}
\mathcal{M}_{pred}^{ans}(L)=
\mathcal{D}\!\left(\mathcal{F}_{img},f_{\phi}\!\left(\mathbf{h}_{seg}^{ans}(L)\right)\right).
\end{equation}
We measure the representation change relative to the zero-length prefix as
\begin{equation}
D_h(L)=1-\cos\!\left(\mathbf{h}_{seg}^{ans}(0),\mathbf{h}_{seg}^{ans}(L)\right).
\label{eq:prefix_hidden_drift}
\end{equation}
Within each model, the image features and decoder remain fixed across the intervention. Changes in $\mathbf{h}_{seg}^{ans}(L)$ can therefore alter the decoder's latent condition without retraining or replacing the segmentation architecture.

\subsection{Controlled Prefix-Length Experiments}
We construct the experiment from the EarthReason test set, using one canonical trajectory saved from GeoReason's naturally generated response for each instance. For a trajectory containing $T$ tokenizer tokens, we retain its first $\lfloor rT\rfloor$ tokens for $r\in\{0,0.25,0.5,0.75,1\}$ and concatenate this prefix with the same final answer. We compare GeoReason with the Qwen3-VL+SAM2 baseline by feeding the resulting sequence into each model. To examine these mechanisms, we measure image attention and hidden-state drift of the answer-side \texttt{[SEG]} across prefix lengths.

\begin{table}[htbp]
\centering
\caption{Effect of reasoning-prefix length on the answer-side \texttt{[SEG]} on EarthReason test. Image-attention mass is summed over image tokens and averaged across heads, the last four layers, and instances. Hidden-state drift is the mean cosine distance between the last-layer \texttt{[SEG]} representations at the retained and zero-length prefixes.}
\label{tab:controlled_prefix_attention}
\small
\begin{tabular*}{0.85\linewidth}{@{\extracolsep{\fill}}rcccc@{}}
\toprule
\multirow{2}{*}{Prefix (\%)} & \multicolumn{2}{c}{Image-attention mass} & \multicolumn{2}{c}{Hidden-state drift} \\
\cmidrule(lr){2-3}\cmidrule(lr){4-5}
& Baseline & GeoReason & Baseline & GeoReason \\
\midrule
0   & 0.2389 & 0.2593 & 0.0000 & 0.0000 \\
25  & 0.2206 & 0.2502 & 0.1955 & 0.0233 \\
50  & 0.2135 & 0.2491 & 0.1896 & 0.0248 \\
75  & 0.2199 & 0.2481 & 0.2097 & 0.0280 \\
100 & 0.2003 & 0.2446 & 0.2076 & 0.0495 \\
\bottomrule
\end{tabular*}
\end{table}

From the zero to the full prefix, the baseline's image attention decreases from $0.2389$ to $0.2003$ ($16.15\%$ relative attenuation), whereas GeoReason changes from $0.2593$ to $0.2446$ ($5.69\%$). GeoReason reduces the absolute endpoint attenuation by $61.7\%$. Its full-prefix hidden-state drift is $0.0495$, compared with $0.2076$ for the baseline, a $76.2\%$ reduction. The baseline's reduced image attention and increased hidden-state drift are consistent with visual-token competition and mask-query drift, respectively. GeoReason mitigates both effects, preserving visual access and stabilizing the segmentation query under longer reasoning prefixes.

\section{Extended Ablation Studies on PTA Loss}

We conduct extensive ablation experiments to validate the design of the \textit{Latent Patch-to-Token Alignment} (PTA) loss, focusing on the PTA head architecture, spatial supervision granularity, pooling strategy, and loss formulations.

\paragraph{Architecture of the PTA Head.} To distinguish from the mask-decoder adapter ($f_\phi$), we denote the auxiliary alignment module as the \textbf{PTA Head ($g_\theta$)}. This head projects the \texttt{[SEG]} token into a dedicated alignment space during training. As shown in Table~\ref{tab:pta_ablation}, removing the PTA head (\textit{w/o PTA Head}) leads to a significant performance drop, especially on EarthReason ($-4.67\%$ in test cIoU). This observation supports our hypothesis that while the \texttt{[SEG]} token and image patch tokens share the same dimensionality, they reside in distinct semantic feature spaces. Directly forcing alignment without a cross-modal projection head (like our GLU-based $g_\theta$) compromises the token's representational capacity, causing it to lose critical information required for subsequent mask generation. Furthermore, our GLU-based PTA head outperforms a simple MLP, effectively modeling complex latent interactions for better alignment.

\paragraph{Granularity of Spatial Supervision.} We compare our mask-based alignment with a coarser bounding box-based strategy. While the BBox approach aligns the \texttt{[SEG]} token with all patches within the rectangular area, our default strategy strictly maps only positive pixels to their corresponding tokens. Table~\ref{tab:pta_ablation} shows that this shape-aware supervision is superior, as it filters out background noise within the BBox and provides more precise geometric guidance.

\paragraph{Pooling Strategy.} We compare our default \textit{Average Pooling}, which uniformly aggregates target patches, with \textit{Attention Pooling} that dynamically weights patches using the \texttt{[SEG]} token as a query. Interestingly, Attention Pooling leads to performance degradation. This is likely because the attention mechanism over-focuses on local salient features while neglecting the overall footprint and boundaries, which are essential for maintaining the structural integrity of masks in complex remote sensing scenes.

\paragraph{Loss Formulation.} Finally, we evaluate different loss functions for PTA. While MSE and MAE provide competitive results, Cosine Similarity achieves the most balanced performance across both benchmarks, demonstrating its robustness in aligning high-dimensional latent vectors.

\begin{table}[htbp]
\centering
\caption{Ablation studies on PTA Loss configurations. $\Delta$ denotes the performance gap relative to our final GeoReason model (red for decrease, blue for increase). Results are reported in (\%).}
\label{tab:pta_ablation}
\resizebox{\textwidth}{!}{
\begin{tabular}{l|cccc|cccc}
\toprule
\multirow{2}{*}{\textbf{Method / Variant}} & \multicolumn{4}{c|}{\textbf{RRSIS-D (Referring)}} & \multicolumn{4}{c}{\textbf{EarthReason (Reasoning)}} \\
\cline{2-9}
& Val gIoU & Val cIoU & Test gIoU & Test cIoU & Val gIoU & Val cIoU & Test gIoU & Test cIoU \\
\midrule
\textbf{GeoReason} & \textbf{68.36} & \textbf{79.87} & \textbf{67.32} & \textbf{79.39} & \textbf{77.61} & \textbf{77.63} & \textbf{78.61} & \textbf{80.60} \\
\midrule
w/o PTA Head & 68.26 \textcolor{red}{\scriptsize{-0.10}} & 78.26 \textcolor{red}{\scriptsize{-1.61}} & 67.24 \textcolor{red}{\scriptsize{-0.08}} & 78.38 \textcolor{red}{\scriptsize{-1.01}} & 77.20 \textcolor{red}{\scriptsize{-0.41}} & 75.92 \textcolor{red}{\scriptsize{-1.71}} & 77.31 \textcolor{red}{\scriptsize{-1.30}} & 75.93 \textcolor{red}{\scriptsize{-4.67}} \\
Simple PTA Head (MLP) & 68.39 \textcolor{blue}{\scriptsize{+0.03}} & 78.85 \textcolor{red}{\scriptsize{-1.02}} & 67.46 \textcolor{blue}{\scriptsize{+0.14}} & 78.80 \textcolor{red}{\scriptsize{-0.59}} & 77.51 \textcolor{red}{\scriptsize{-0.10}} & 78.60 \textcolor{blue}{\scriptsize{+0.97}} & 78.34 \textcolor{red}{\scriptsize{-0.27}} & 79.78 \textcolor{red}{\scriptsize{-0.82}} \\
\midrule
BBox-based Alignment & 68.56 \textcolor{blue}{\scriptsize{+0.20}} & 79.26 \textcolor{red}{\scriptsize{-0.61}} & 67.28 \textcolor{red}{\scriptsize{-0.04}} & 78.65 \textcolor{red}{\scriptsize{-0.74}} & 77.26 \textcolor{red}{\scriptsize{-0.35}} & 77.71 \textcolor{blue}{\scriptsize{+0.08}} & 78.64 \textcolor{blue}{\scriptsize{+0.03}} & 79.47 \textcolor{red}{\scriptsize{-1.13}} \\

PTA Pooling: Attention & 67.76 \textcolor{red}{\scriptsize{-0.60}} & 79.05 \textcolor{red}{\scriptsize{-0.82}} & 66.69 \textcolor{red}{\scriptsize{-0.63}} & 78.11 \textcolor{red}{\scriptsize{-1.28}} & 77.47 \textcolor{red}{\scriptsize{-0.14}} & 77.47 \textcolor{red}{\scriptsize{-0.16}} & 78.10 \textcolor{red}{\scriptsize{-0.51}} & 78.65 \textcolor{red}{\scriptsize{-1.95}} \\
\midrule
PTA Loss: MSE & 68.69 \textcolor{blue}{\scriptsize{+0.33}} & 79.45 \textcolor{red}{\scriptsize{-0.42}} & 67.12 \textcolor{red}{\scriptsize{-0.20}} & 78.30 \textcolor{red}{\scriptsize{-1.09}} & 77.40 \textcolor{red}{\scriptsize{-0.21}} & 77.06 \textcolor{red}{\scriptsize{-0.57}} & 78.49 \textcolor{red}{\scriptsize{-0.12}} & 79.02 \textcolor{red}{\scriptsize{-1.58}} \\
PTA Loss: MAE & 68.27 \textcolor{red}{\scriptsize{-0.09}} & 79.60 \textcolor{red}{\scriptsize{-0.27}} & 67.23 \textcolor{red}{\scriptsize{-0.09}} & 78.61 \textcolor{red}{\scriptsize{-0.78}} & 77.54 \textcolor{red}{\scriptsize{-0.07}} & 77.22 \textcolor{red}{\scriptsize{-0.41}} & 78.66 \textcolor{blue}{\scriptsize{+0.05}} & 78.40 \textcolor{red}{\scriptsize{-2.20}} \\
\bottomrule
\end{tabular}
}
\end{table}

\section{Additional Benchmark Results on GeoSeg-Bench}
\label{appendix:geoseg_transfer}

GeoSeg-Bench~\cite{ni2025unigeoseg} is a geospatial segmentation benchmark encompassing Interactive, Referring, and Reasoning tasks across diverse scenes and categories. To further demonstrate GeoReason's generalizability across diverse and challenging geospatial contexts, we extend our evaluation to this benchmark. We evaluate the GeoReason checkpoint trained in the main experiments on the Referring and Reasoning subsets of GeoSeg-Bench, without fine-tuning on GeoSeg-1M, the training dataset introduced in UniGeoSeg~\cite{ni2025unigeoseg}. Table~\ref{tab:geoseg_direct_transfer} compares GeoReason with eleven methods evaluated without GeoSeg-1M fine-tuning, as well as fine-tuned Geopixel and Earthmind, using results reported in UniGeoSeg~\cite{ni2025unigeoseg}.

\begin{table}[tb]
\centering
\caption{\small{Results on GeoSeg-Bench. FT indicates fine-tuning on the GeoSeg-1M training dataset~\cite{ni2025unigeoseg}. Baseline scores are reproduced from UniGeoSeg~\cite{ni2025unigeoseg}. Best results are in \textbf{bold}; second-best are \underline{underlined}.}}
\label{tab:geoseg_direct_transfer}
\small
\begin{tabular*}{0.85\linewidth}{@{\extracolsep{\fill}}llcccc@{}}
\toprule
\multirow{2}{*}{Method} & \multirow{2}{*}{FT} & \multicolumn{2}{c}{Referring} & \multicolumn{2}{c}{Reasoning} \\
\cmidrule(lr){3-4}\cmidrule(lr){5-6}
& & gIoU & cIoU & gIoU & cIoU \\
\midrule
LISA~\cite{lisa} & \xmark & 4.56  & 3.53  & 5.77  & 7.09 \\
PixelLM~\cite{ren2024pixellm} & \xmark & 5.80  & 6.04  & 6.30  & 6.36 \\
PSALM~\cite{zhang2024psalm} & \xmark & 18.91 & 31.77 & 9.27  & 11.88 \\
HIPIE~\cite{wang2023hipie} & \xmark & 39.14 & 29.25 & 11.76 & 7.90 \\
SegLLM~\cite{wang2024segllm} & \xmark & 35.27 & 15.90 & 16.41 & 11.06 \\
Geopixel~\cite{shabbir2025geopixel} & \xmark & 40.14 & 37.34 & 26.71 & 27.36 \\
Geopix~\cite{ou2025geopix} & \xmark & 28.52 & 28.81 & 18.25 & 20.80 \\
RemoteSAM~\cite{yao2025remotesam} & \xmark & 29.31 & 12.06 & 9.01  & 8.09 \\
Earthmind~\cite{shu2025earthmind} & \xmark & 46.48 & 44.53 & \underline{28.80} & \underline{31.01} \\
LISAT~\cite{quenum2025lisat} & \xmark & 40.14 & 37.87 & 20.09 & 22.06 \\
SegEarth-R1~\cite{segearthr1} & \xmark & 9.39  & 14.65 & 11.97 & 12.72 \\
\midrule
Geopixel~\cite{shabbir2025geopixel} & \cmark & 45.50 & 42.70 & 28.51 & 30.25 \\
Earthmind~\cite{shu2025earthmind} & \cmark & \underline{49.24} & \underline{48.09} & 25.71 & \textbf{36.84} \\
\midrule
\textbf{GeoReason} & \xmark & \textbf{56.64} & \textbf{50.50} & \textbf{32.46} & 30.99 \\
\bottomrule
\end{tabular*}
\end{table}

As shown in Table~\ref{tab:geoseg_direct_transfer}, GeoReason achieves the highest gIoU on both tasks among the compared methods. Compared with Earthmind, the next-best method without GeoSeg-1M fine-tuning, GeoReason improves Referring and Reasoning gIoU by $10.16$ and $3.66$ percentage points, respectively. Referring cIoU improves by $5.97$ points, while Reasoning cIoU is nearly tied ($30.99$ versus $31.01$, a difference of $-0.02$ points). Despite using no GeoSeg-1M training data, GeoReason also surpasses the fine-tuned baselines on multiple metrics. These results further demonstrate its effectiveness beyond EarthReason and RRSIS-D.

\section{Qualitative Visualizations and Analysis}
\label{sec:qualitative}

In this section, we provide detailed qualitative evidence to further elucidate how GeoReason addresses the challenges in remote sensing (RS) reasoning segmentation.

\subsection{Case Study}
\label{subsec:case_study}
\begin{figure}[!htbp]
    \centering
    \includegraphics[width=1\linewidth]{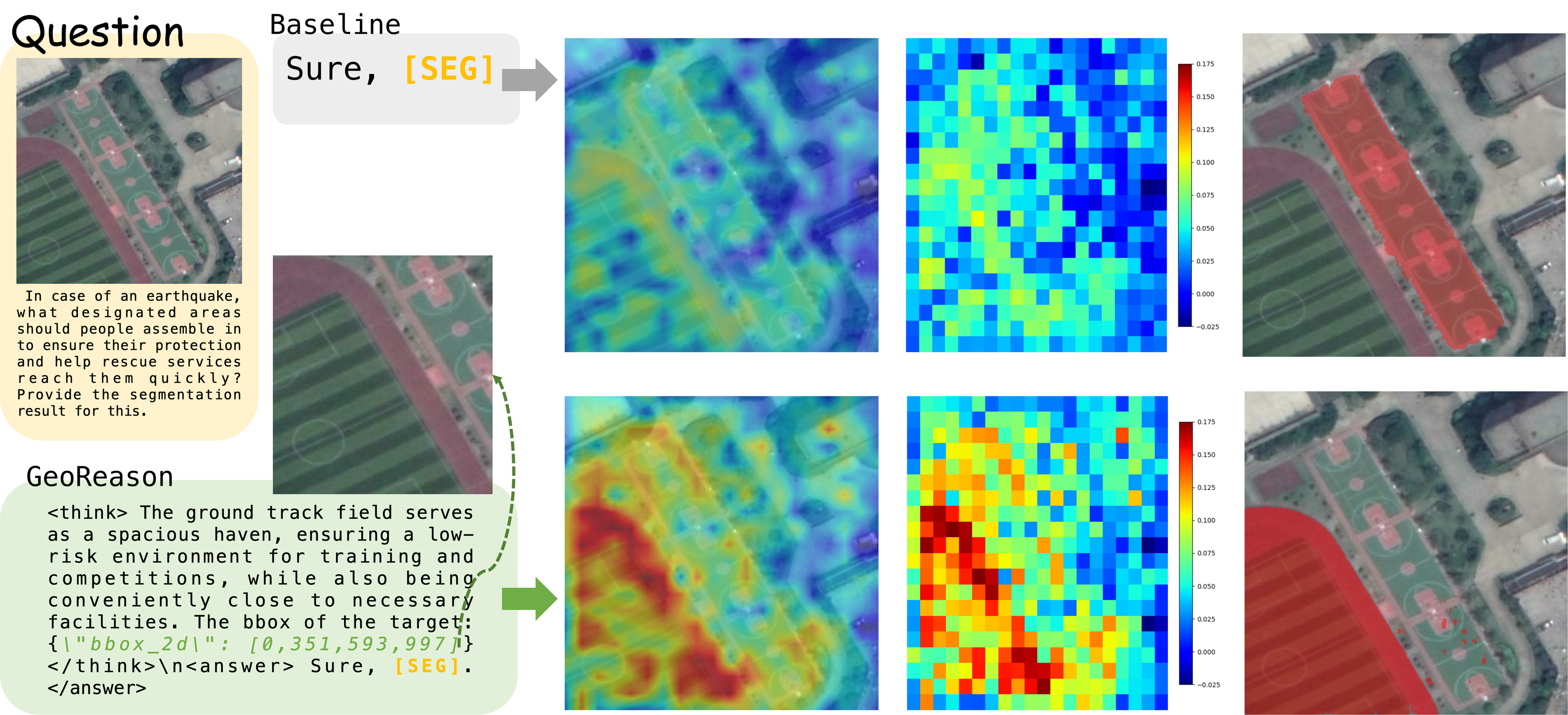}
\caption{Qualitative case study comparing the baseline and GeoReason. GeoReason first grounds the correct region with an intermediate bbox in CoT, then exhibits stronger target-focused \texttt{[SEG]} and produces a more accurate final mask.}
    \label{fig:case_study}
\end{figure}
We further illustrate GeoReason with a representative case study against the baseline (Qwen3VL-4B + SAM2 without our proposed components). As shown in Fig.~\ref{fig:case_study}, the baseline attends to an incorrect structure and fails to isolate the intended target, whereas GeoReason correctly identifies the target region. A key difference is that GeoReason first localizes the relevant area through the intermediate bbox generated in CoT, which provides an explicit coarse spatial anchor before mask decoding.

We then visualize the cosine similarity between the \texttt{[SEG]} hidden state and image patch tokens. Compared with the baseline, GeoReason produces a much stronger and more concentrated response on the true target region, while suppressing unrelated areas. This sharper contrast is consistent with the final segmentation result: GeoReason yields the correct mask, whereas the baseline remains distracted by competing regions and produces an inferior prediction.

This case supports the complementary roles of our components. The Location-aware CoT improves coarse target localization through bbox reasoning, while the Anticipatory Prior and PTA loss make the \texttt{[SEG]} representation more target-sensitive, leading to more reliable final segmentation.

\subsection{Analysis of Typical Failure Modes}
\label{subsec:failure_modes}
\begin{figure}[!htbp]
    \centering
    \includegraphics[width=1\linewidth]{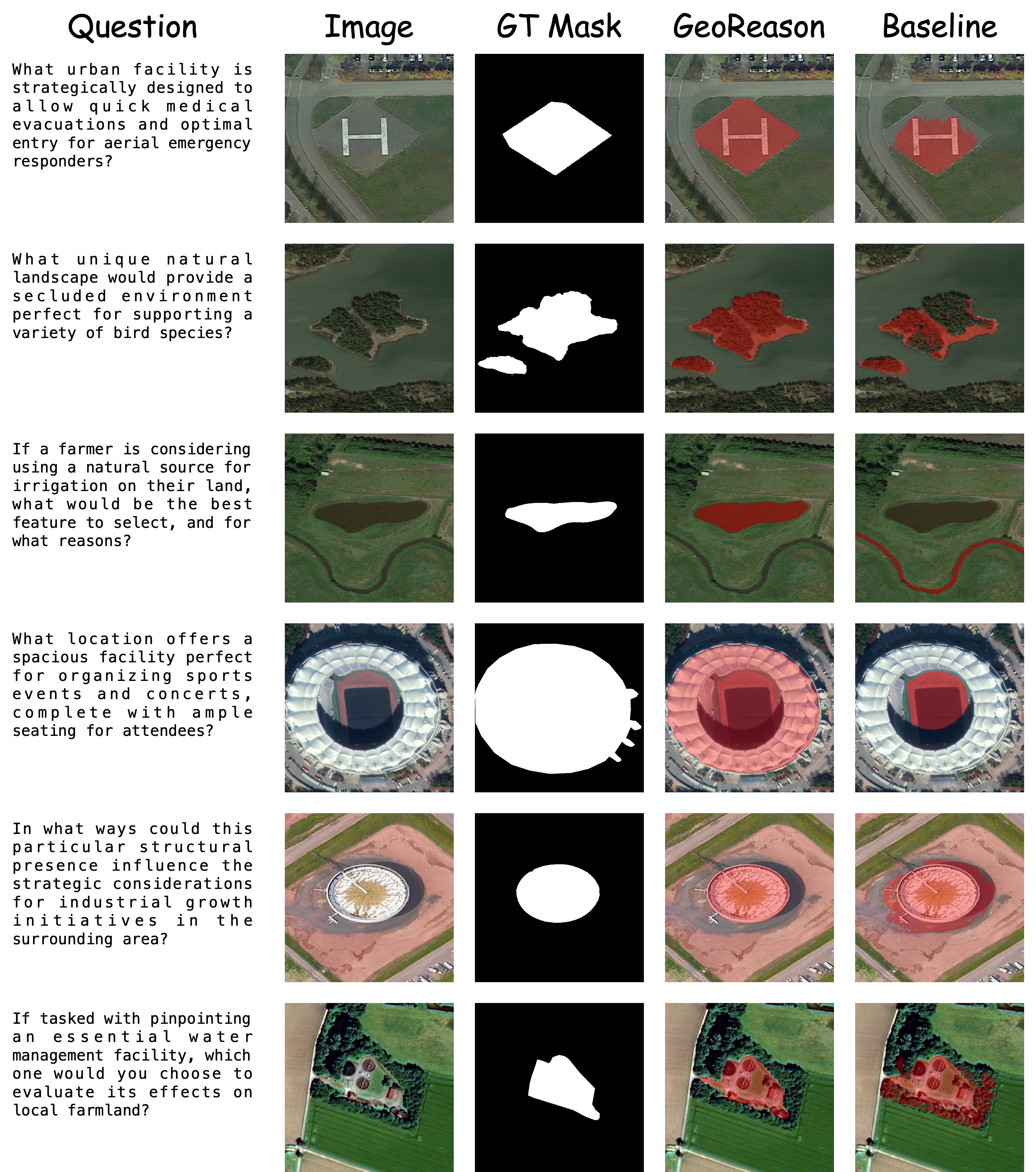}
\caption{Mitigation of typical failure modes in RS reasoning segmentation.
}
    \label{fig:more_visuals}
\end{figure}
We further provide additional qualitative results to illustrate how GeoReason addresses common failure modes in remote sensing reasoning segmentation (Fig. \ref{fig:more_visuals}):

\begin{enumerate}
    \item \textbf{Spatial Fidelity Degradation:} Due to information decay, baseline models often generate noisy or fragmented masks with blurry boundaries. GeoReason establishes a spatial anchor on the prompt-side via the \textit{Anticipatory Prior}, anchoring visual evidence before autoregressive generation to maintain mask integrity.
    
    \item \textbf{Target Misidentification:} Weak visual reasoning often leads standard models to segment incorrect objects or partial targets. Our \textit{Location-aware CoT} enforces an explicit coarse localization (bounding box) before generating the \texttt{[SEG]} token, ensuring the model focuses on the correct location.
    
    \item \textbf{Boundary Redundancy:} Insufficient understanding of RS textures causes baseline masks to exhibit redundant fragments at the boundaries. By applying the \textit{PTA Loss}, we align the \texttt{[SEG]} token with ground-truth image patches, enhancing its representational quality and eliminating blurry mask redundancies.
\end{enumerate}

\FloatBarrier
\section{Limitations}
Our study has several limitations. First, the main experiments use Qwen-VL and SAM-family models. Although the proposed design is modular, its transferability to other MLLM backbones or segmentation decoders has not yet been systematically evaluated. Second, our benchmarks mainly emphasize single-target segmentation, and more complex settings such as multi-target reasoning, heavy occlusion, or highly compositional queries remain for future study.

\end{document}